%% file: main.tex
\documentclass[10pt,twocolumn,letterpaper]{article}

\usepackage[pagenumbers]{cvpr}

\input{preamble}

\definecolor{cvprblue}{rgb}{0.21,0.49,0.74}
\usepackage[breaklinks,colorlinks,allcolors=cvprblue]{hyperref}
\hypersetup{pdfauthor={Zhiming Xu, Huiyu Yi, Baile Xu, Furao Shen, Jian Zhao, Suorong Yang},
  pdftitle={From Class-Centric to Joint Class-Property Modeling in CLIP-Based Continual Learning}}

\def\confName{CVPR}
\def\confYear{2027}

\title{From Class-Centric to Joint Class-Property Modeling \\ in CLIP-Based Continual Learning}

\author{
    Zhiming Xu\textsuperscript{1,2},
    Huiyu Yi\textsuperscript{1,2},
    Baile Xu\textsuperscript{1,2},
    Furao Shen\textsuperscript{1,2,\textdagger},
    Jian Zhao\textsuperscript{4},
    Suorong Yang\textsuperscript{1,3}
}

\newcommand{\affiliationfootnote}[1]{%
  \begingroup
  \renewcommand{\thefootnote}{}%
  \footnotetext[0]{#1}%
  \endgroup
}

\begin{document}
\maketitle
\affiliationfootnote{%
\textsuperscript{1}National Key Laboratory for Novel Software Technology, Nanjing University, Nanjing, China.\\
\textsuperscript{2}School of Artificial Intelligence, Nanjing University, Nanjing, China.\\
\textsuperscript{3}School of Computer Science, Nanjing University, Nanjing, China.\\
\textsuperscript{4}School of Electronic Science and Engineering, Nanjing University, China.\\
\textsuperscript{\textdagger}Correspondence to: \texttt{frshen@nju.edu.cn}.
}
\input{sec/abstract}    
\input{sec/intro}
\input{sec/related}

\input{sec/pre}

\input{sec/method}

\input{sec/exp}
\input{sec/conclu}

{
    \small
    \bibliographystyle{ieeenat_fullname}
    \bibliography{main}
}


\end{document}

%% file: preamble.tex
\usepackage{booktabs,tabularx,array,multirow}
\newcolumntype{Y}{>{\centering\arraybackslash}X}



%% file: sec/abstract.tex
\begin{abstract}
Although CLIP-based continual learning has achieved promising performance, existing methods primarily focus on continual class prediction from inputs, largely overlooking the explicit prediction of properties that characterize objects at different granularities.
Such properties provide complementary cues for class discrimination, but exploiting them in continual learning requires reliable property prediction and effective preservation of previously acquired property knowledge as new classes arrive. We propose Separated Training with Reliability-Aware Interaction for Dual Estimation (STRIDE), a framework that jointly performs classification and property prediction in continual learning. 
STRIDE introduces a property prediction pathway that models object properties from global and local features. Property-derived evidence both refines the representations used for base classification and feeds a separate property-to-class classifier, producing complementary class logits that combine with the base logits for final prediction. To reduce interference between pathway establishment and property-guided classification, STRIDE adopts two-stage learning: Stage A independently establishes the classification and property prediction pathways, while Stage B optimizes property-guided representation enhancement and property-to-class prediction.
Extensive experiments across multiple continual learning benchmarks demonstrate that STRIDE achieves superior classification performance while maintaining strong property prediction performance.

\end{abstract}

%% file: sec/intro.tex
\section{Introduction}
\label{sec:intro}
Continual learning (CL) aims to enable models to acquire knowledge from sequentially arriving data \cite{chen2024promptfusion,zheng2025tagfex}. However, adapting to evolving data distributions with limited access to historical samples can compromise previously acquired knowledge, leading to catastrophic forgetting. Recent advances \cite{gao2024cprompt,wang2025sema,gao2025moal,wang2025tuna} in continual learning have been further driven by pre-trained models (PTMs), whose transferable representations provide a strong foundation for incremental adaptation. Building on this advantage, recent methods have achieved promising progress in acquiring new knowledge while mitigating catastrophic forgetting.
\begin{figure}[!t]
  \centering
  \includegraphics[width=2.3in,keepaspectratio]{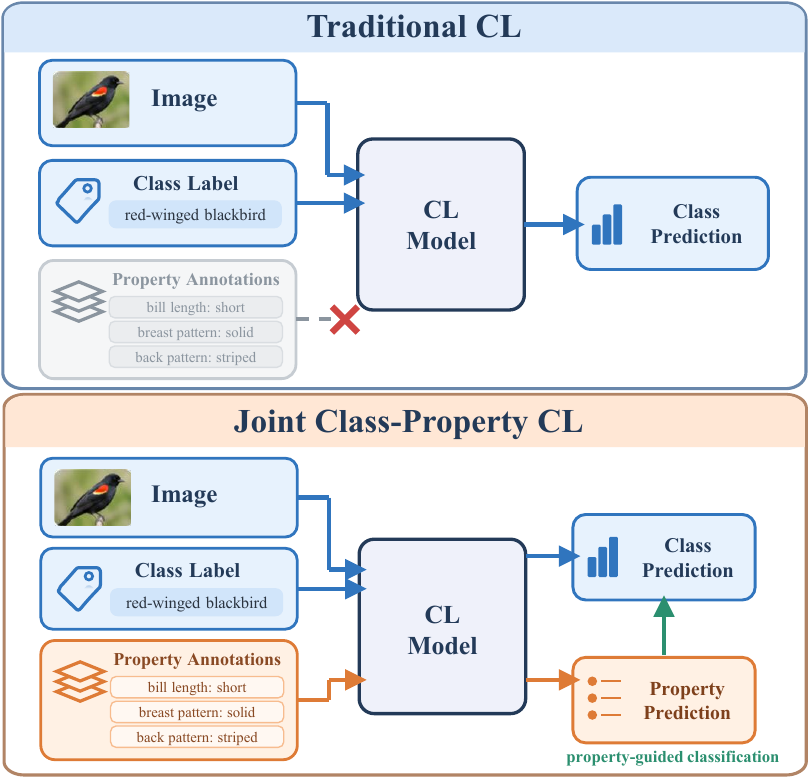}
  \caption{Joint class-property continual learning. Existing methods achieve CL over classes alone. We study CL of both classes and descriptive properties, and explore how property information can improve classification.}
  \label{fig:fig1}
\end{figure}

Vision-language models (VLMs) have emerged as a promising foundation for continual learning by connecting visual understanding with natural language. Among them, CLIP \cite{radford2021clip}, a widely adopted \cite{zhang2025adaptive} VLM, learns transferable visual and textual representations through large-scale image-text contrastive pre-training. Its shared embedding space enables open-vocabulary recognition, image-text retrieval, and generalization to novel visual concepts. These capabilities have motivated growing research on adapting VLMs to continual learning \cite{wang2023attriclip,huang2024rapf,yu2024moeadapters,luo2025lada,huang2025mgclip}, aiming to accommodate newly introduced classes while preserving their rich cross-modal knowledge.

Beyond class labels, many datasets provide structured property annotations that characterize objects from different perspectives \cite{maji2013aircraft,wah2011cub}, ranging from broader category groupings to specific visual characteristics. However, the joint prediction of classes and dataset-provided properties remains largely unexplored in CLIP-based continual learning. Existing methods primarily exploit external textual knowledge or latent attributes to enhance classification \cite{zhou2025engine,xie2026area}, without explicitly learning to predict these properties. Given CLIP's powerful image-text alignment capabilities, property annotations can be naturally represented as textual descriptions, offering the potential to learn classification and property prediction within a unified framework. Accurate property prediction not only provides a more comprehensive understanding of visual content beyond class labels but also offers complementary evidence for distinguishing similar classes. Nevertheless, such classification benefits depend on the reliability of predicted properties, as inaccurate predictions may introduce misleading information. This raises a new question: \textit{How can we jointly achieve accurate classification and property prediction while reliably leveraging predicted properties to further improve classification throughout continual learning?}

To address this challenge, we propose Separated Training with Reliability-Aware Interaction for Dual Estimation (STRIDE), a property-aware dual-prediction framework for CLIP-based continual learning. STRIDE establishes complementary classification and property prediction pathways based on frozen CLIP representations. Since certain properties depend on localized details that may be overlooked by global representations, we further enhance the property prediction pathway with local features, which refine global property estimates through residual analytic prediction. 
To effectively coordinate the two prediction objectives, we develop a decoupled two-stage learning strategy that separates foundational class-property learning from property-guided classification enhancement. The first stage establishes the classification and property prediction pathways through independent learning objectives, while the second stage exploits the learned property evidence to enhance classification under a constrained optimization scheme, preserving the knowledge acquired by both pathways.
Furthermore, STRIDE leverages predicted property evidence at both the representation and prediction levels. Property-guided representation learning enhances class discrimination by incorporating property information into visual features, while an explicit property-to-class pathway provides complementary classification evidence. 
The contribution of this evidence is adaptively controlled according to estimated property confidence and classification uncertainty, improving classification while limiting the influence of uncertain property evidence.
Extensive experiments across multiple benchmarks demonstrate that STRIDE consistently outperforms existing state-of-the-art methods in both classification accuracy and property prediction performance.
The main contributions are summarized as follows: 
\begin{itemize}
    \item We propose STRIDE, a property-aware dual-prediction framework for CLIP-based continual learning. Its property pathway leverages complementary global and local visual features, while a decoupled two-stage strategy separates predictor establishment from property-guided classification refinement to reduce learning interference.
    \item We develop a property-guided classification enhancement mechanism that leverages property evidence to refine representations in the classification pathway and uses a property-to-class classifier to map the same evidence into complementary class logits, which are combined with the base class logits for final prediction.
    \item Extensive experiments across multiple continual learning settings demonstrate that our approach consistently outperforms strong baselines and state-of-the-art methods in both classification and property prediction.
\end{itemize}

%% file: sec/related.tex
\section{Related Work}
\label{sec:related}
\subsection{Continual Learning} 

Continual learning (CL) aims to acquire knowledge from sequentially arriving tasks while preserving previously learned capabilities. Replay-based methods mitigate catastrophic forgetting by improving memory utilization and stabilizing optimization with historical samples \cite{lin2023pcr,yoo2024lpr}. Distillation-based approaches preserve historical knowledge through relational constraints or complementary teacher models \cite{asadi2023psrd,wen2024mtd}. Other methods explore exemplar compression \cite{luo2023ciec}, feature calibration and saliency preservation \cite{li2024fcs,liu2024tass}, and dynamic feature expansion \cite{hu2023dne,zheng2025tagfex}. More recently, pre-trained models (PTMs) have significantly advanced continual learning through diverse strategies, including prompt tuning \cite{smith2023coda,wang2022dualprompt}, interference-aware subspace adaptation \cite{he2025cllora}, architectural expansion \cite{zhou2024ease,li2025dia}, classifier alignment and analytic learning \cite{zhang2023slca,mcdonnell2023ranpac}, and parameter adaptation \cite{wang2025tuna, zhao2026rsiat}.

\subsection{CLIP-Based Continual Learning}
Recent studies have explored adapting CLIP to continual learning by leveraging its cross-modal knowledge. RAPF employs adaptive representation adjustment and parameter fusion to mitigate forgetting \cite{huang2024rapf}, while PROOF \cite{zhou2025learning} introduces task-specific projections and cross-modal feature fusion to integrate visual and textual knowledge. Other approaches explore parameter-efficient adaptation \cite{yu2024moeadapters,luo2025lada}, cross-modal representation alignment \cite{huang2025mgclip,li2026bofa}, and synthetic data distillation \cite{wu2025gift}. Beyond model adaptation, recent methods incorporate additional knowledge into continual classification. CLG-CBM \cite{yu2025clgcbm} employs language-guided concepts to improve interpretability and classification. ENGINE enriches class representations with external textual knowledge generated by GPT-4 \cite{zhou2025engine}. More recently, AREA \cite{xie2026area} extracts and aggregates latent class-level attributes to support continual classification. In contrast, STRIDE explicitly learns to predict dataset-annotated text properties as an independent objective throughout continual learning. These predicted properties are further leveraged to enhance visual representations and provide complementary classification evidence through a property-to-class pathway.

%% file: sec/pre.tex
\begin{figure*}[!ht]
  \centering
  \includegraphics[width=5in,keepaspectratio]{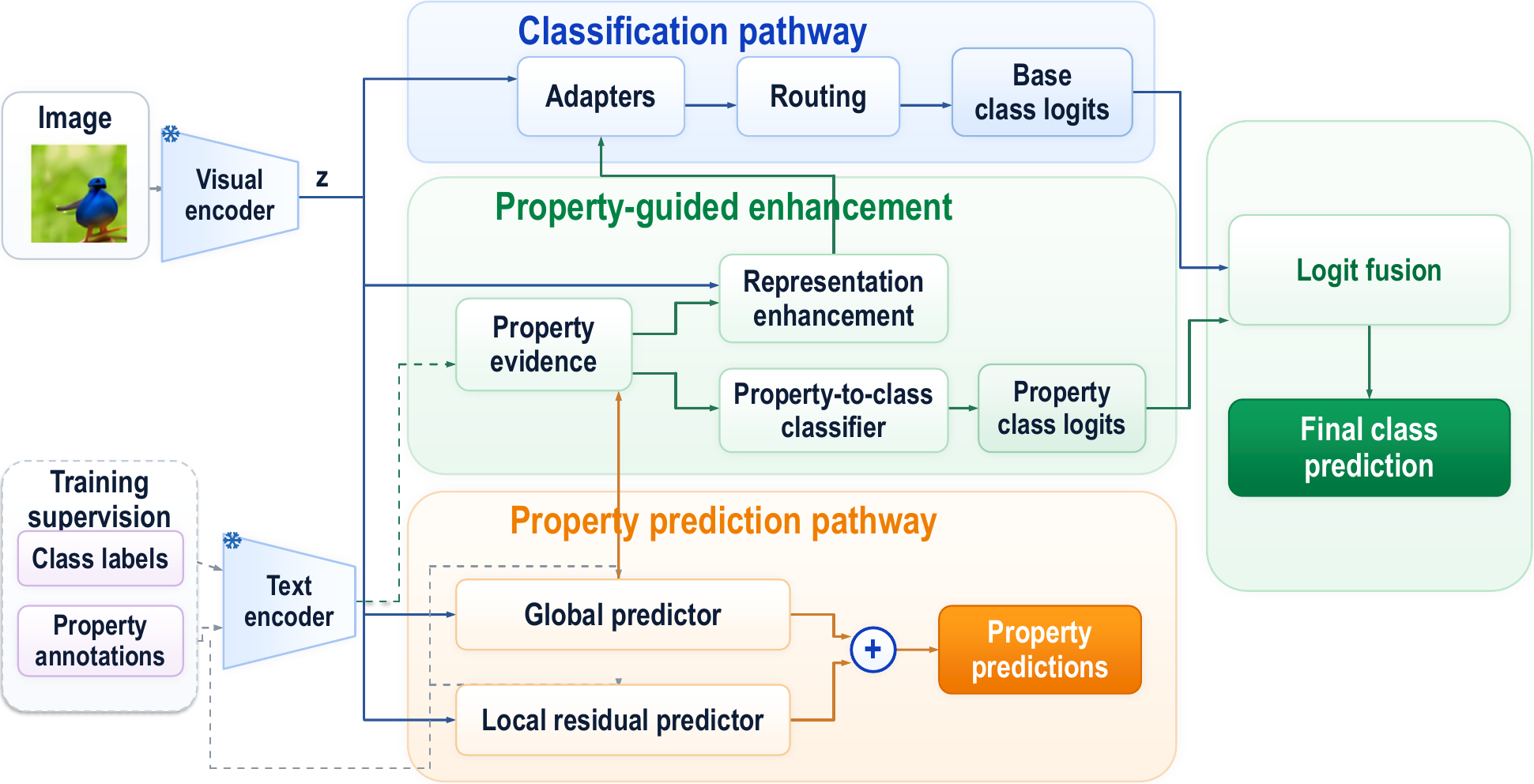}
  \caption{Illustration of STRIDE. Given an input image, STRIDE jointly predicts its class and properties from visual features. Property evidence derived from the global predictor guides visual representation enhancement along the classification pathway, while a separate property-to-class classifier maps the property evidence into property class logits. The enhanced visual features yield base class logits, which are adaptively fused with property class logits to produce the final class prediction.}
    \label{fig:ours}
\end{figure*}

\section{Preliminaries}

\subsection{Continual Learning with CLIP}
Continual learning (CL) aims to acquire knowledge from sequentially arriving data while preserving previously learned capabilities \cite{ijcai2024p0924}. Under the widely adopted exemplar-free class-incremental learning setting \cite{zhou2024class,zhou2024ease}, learning proceeds over $T$ tasks $\{\mathcal{D}_t\}_{t=1}^{T}$, each introducing a disjoint set of classes $\mathcal{C}_t$. Only the current task data $\mathcal{D}_t$ are accessible during training, without retaining historical training samples \cite{xie2026area}. At inference, the model receives only a test image $x$ and must classify it among all observed classes $\mathcal{C}_{1:t}=\bigcup_{j=1}^{t}\mathcal{C}_j$, without knowing its task identity. CLIP \cite{radford2021clip} consists of a visual encoder $E_v$ and a text encoder $E_t$ that map images and texts into a shared embedding space. In its basic classification mechanism, textual class representations $\mathbf{w}_c=E_t(s_c)$ are constructed from pre-defined class prompts $s_c$. Given only the test image $x$, CLIP extracts its visual representation and computes classification probabilities by matching it with the class text representations:
\begin{equation}
P(y=c\mid x)=
\frac{\exp\left(\mathrm{sim}(E_v(x),\mathbf{w}_c)/\tau\right)}
{\sum_{j\in\mathcal{C}_{1:t}}\exp\left(\mathrm{sim}(E_v(x),\mathbf{w}_j)/\tau\right)},
\end{equation}
where $\mathrm{sim}(\cdot,\cdot)$ denotes cosine similarity and $\tau$ is a temperature parameter. Existing methods further introduce learnable adaptation modules or additional classifiers to improve incremental classification \cite{zhou2025learning,zhou2025engine,xie2026area}.

\subsection{Class-Property Dual Prediction in CL}
Beyond conventional classification, we extend the learning objective to include explicit property prediction. Specifically, each training sample is represented as $(x_i, y_i, \mathbf{p}_i)$, where $x_i$ denotes an image, $y_i$ is its class label, and $\mathbf{p}_i=(p_i^{(1)},\ldots,p_i^{(K)})$ contains labels for $K$ predefined property dimensions. While these dimensions remain fixed, their possible values are incrementally observed as new classes arrive. We denote by $\mathcal{P}_k^{(t)}$ the set of observed candidate values for the $k$-th property dimension up to task $t$. Each property value is represented as a textual description that can be encoded by CLIP's text encoder, enabling its learning through image-text alignment. At each task $t$, the model is required to jointly predict the class and properties of an input image:
\begin{equation}
\hat{y}_t=f_t^{\mathrm{cls}}(x), \qquad
\hat{\mathbf{p}}_t=f_t^{\mathrm{prop}}(x),
\end{equation}
where $\hat{y}_t\in\mathcal{C}_{1:t}$ and $\hat{\mathbf p}_t\in\prod_{k=1}^{K}\mathcal{P}_k^{(t)}$. During inference, only the image is provided, without access to its class label, property annotations, or task ID. The objective is to acquire new class and property knowledge while preserving previously learned predictions, and to leverage property information to further improve classification.

%% file: sec/method.tex
\section{Method}

\subsection{Class-Property Dual Pathway}
Following existing CLIP-based CL methods \cite{zhou2025learning,xie2026area}, STRIDE employs a pre-trained CLIP backbone with visual and textual encoders $E_v$ and $E_t$, which map images and texts into a shared embedding space. Both encoders remain frozen throughout learning.
Given an image $x$ and a textual description $s$, the encoders extract representations $\mathbf{z}=E_v(x)\in\mathbb{R}^{d}$ and $\mathbf{t}=E_t(s)\in\mathbb{R}^{d}$, respectively, which are fed into two pathways for classification and property prediction. 

We adopt established architectures \cite{zhou2025learning,zhou2025engine,xie2026area} for the base classification pathway. Specifically, task-shared visual and textual adapters refine the frozen CLIP representations, while task-specific scoring modules, each equipped with learnable visual and textual projections, are incrementally introduced to distinguish newly acquired classes. During inference, the visual and textual adapters transform the CLIP representations into $\tilde{\mathbf{z}}=\mathcal{A}_v(\mathbf{z})$ and $\tilde{\mathbf{T}}=\mathcal{A}_t(\mathbf{T})$, where $\mathbf{T}$ contains the textual representations of all observed classes, and $\mathcal{A}_v$ and $\mathcal{A}_t$ denote the visual and textual adapters, respectively. The adapted features are further combined with scoring modules to compute classification scores, which are aggregated through task-agnostic routing and integrated with Gaussian Discriminant Analysis (GDA) \cite{wang2024baseline,xie2026area}:
\begin{equation}
\mathbf{s}_{\mathrm{cls}}(x)
=
(1-\lambda_g)\sum_{j=1}^{t}
\pi_j(x)\mathbf{s}_{j}(\tilde{\mathbf{z}},\tilde{\mathbf{T}})
+
\lambda_g\mathbf{s}_{\mathrm{GDA}}(x),
\label{eq:base_classification}
\end{equation}
where $\mathbf{s}_{\mathrm{cls}}(x)$ denotes the class prediction logits, $\mathbf{s}_j$ denotes the $j$-th scoring module, $\pi_j(x)$ is its routing weight, and $\lambda_g$ controls the contribution of GDA.

For the property prediction pathway, STRIDE considers the spatially localized nature of visual properties. Unlike class labels, which can often be inferred from overall appearance, certain properties depend on specific regions, such as a bird's breast pattern or bill shape. These details may be underrepresented in global CLIP features. To capture such information, STRIDE uses attention maps from the frozen visual encoder to locate informative regions, which are cropped and re-encoded to extract local features. These features then refine global property predictions through a residual formulation:
\begin{equation}
\mathbf{q}_k(x)=
\operatorname{softmax}\left(
g_k(\mathbf{z})+
r_k(\mathbf{z},\mathbf{z}_k^{\mathrm{loc}})
\right),
\label{eq:property_prediction}
\end{equation}
where $\mathbf{q}_k(x)$ denotes the predicted probability distribution over $\mathcal{P}_k^{(t)}$ for the $k$-th property dimension and $\mathbf{z}_k^{\mathrm{loc}}$ represents the local visual features extracted from property-relevant regions. The functions $g_k$ and $r_k$ are ridge regression-based linear predictors \cite{hoerl1970ridge,mcdonnell2023ranpac} that produce property logits, with $g_k$ estimating global logits and $r_k$ providing local residual corrections. Their parameters are obtained through regularized least squares, enabling efficient incremental updates without gradient-based optimization.

\subsection{Property-Guided Classification Enhancement}
STRIDE further exploits property estimates to enhance classification at both the representation and prediction levels. We construct classification evidence from the global predictor $g_k(z)$, which operates on the same full-image features as the classification pathway. To represent this evidence in the CLIP embedding space, we use the observed property values $v\in\mathcal{P}_k^{(t)}$ as textual prototypes and encode their descriptions with the frozen CLIP text encoder. The resulting text embeddings are then weighted by the Softmax-normalized global property logits to obtain:
\begin{equation}
\mathbf{e}_k(x)=
\sum_{v\in\mathcal{P}_k^{(t)}}
\left[\operatorname{softmax}(g_k(\mathbf{z}))\right]_v
E_t(s_{k,v}),
\label{eq:property_evidence}
\end{equation}
where $s_{k,v}$ denotes the textual description of the property value $v$ in dimension $k$. $\mathbf{e}_k(x)$ are normalized and aggregated according to their prediction confidence to obtain a unified evidence vector:
\begin{equation}
\mathbf{e}(x)=
\mathcal{N}\left(
\sum_{k=1}^{K}
\max_{v\in\mathcal{P}_k^{(t)}}
\left[\operatorname{softmax}(g_k(\mathbf{z}))\right]_v
\mathcal{N}(\mathbf{e}_k(x))
\right),
\label{eq:aggregated_property_evidence}
\end{equation}
where $\mathcal{N}(\cdot)$ denotes L2 normalization. 
The max operation selects the highest predicted probability within each property dimension as its confidence weight, allowing higher-confidence property evidence to contribute more strongly to the aggregated representation.
At the representation level, STRIDE employs an attention-based interaction network $\mathcal{F}_{\mathrm{att}}$ to integrate property evidence with visual features, producing an enhanced representation $\mathbf{z}_{\mathrm{enh}}=\mathcal{F}_{\mathrm{att}}(\mathbf{z},\mathbf{e}(x))$. The enhanced features are subsequently fed into the classification pathway to compute class logits in Eq.~\eqref{eq:base_classification}.
At the prediction level, the aggregated property evidence $\mathbf{e}(x)$ captures predicted properties but is not explicitly optimized for class discrimination. STRIDE therefore employs a learnable residual projection $\mathcal{M}_{\mathrm{pc}}$ to model class-relevant interactions among properties, while retaining a direct pathway for the original evidence. The refined evidence is mapped into complementary class logits through a cosine classifier $\mathcal{C}_{\mathrm{pc}}$:
\begin{equation}
\mathbf{s}_{\mathrm{prop}}(x)
=
\mathcal{C}_{\mathrm{pc}}\big(
\mathbf{e}(x)+\mathcal{M}_{\mathrm{pc}}(\mathbf{e}(x))
\big).
\label{eq:property_class_logits}
\end{equation}
To integrate property-based predictions with the base classification pathway, we align their logit scales and adaptively adjust the property contribution according to property confidence and classification uncertainty:
\begin{equation}
\mathbf{s}_{\mathrm{final}}(x)
=
\mathbf{s}_{\mathrm{cls}}(x)
+
\alpha(x)
\frac{\sigma_{\mathrm{cls}}}{\sigma_{\mathrm{prop}}}
\left(
\mathbf{s}_{\mathrm{prop}}(x)-\mu_{\mathrm{prop}}
\right),
\label{eq:property_class_fusion}
\end{equation}
where $\mathbf{s}_{\mathrm{cls}}(x)$ denotes the classification logits computed using $\mathbf{z}_{\mathrm{enh}}$. Specifically, $\mu_{\mathrm{prop}}$ and $\sigma_{\mathrm{prop}}$ are the mean and standard deviation of $\mathbf{s}_{\mathrm{prop}}(x)$ across all observed classes, while $\sigma_{\mathrm{cls}}$ is the standard deviation of $\mathbf{s}_{\mathrm{cls}}(x)$. Subtracting $\mu_{\mathrm{prop}}$ removes the bias in property-based logits, while $\sigma_{\mathrm{cls}}/\sigma_{\mathrm{prop}}$ adjusts their magnitude to match the variability of the classification logits. This prevents differences in logit scales from disproportionately influencing the final prediction. The adaptive coefficient $\alpha$ is defined as:

\begin{equation}
\alpha(x)=\operatorname{clip}\left(
\alpha_{\min}+\beta\,c(x)u(x),\,0,\,1
\right),
\label{eq:adaptive_property_fusion}
\end{equation}
where $c(x)=\frac{1}{K}\sum_{k=1}^{K}\max_{v\in\mathcal{P}_k^{(t)}}[\operatorname{softmax}(g_k(\mathbf{z}))]_v$ measures the average property prediction confidence, and $u(x)=H(\operatorname{softmax}(\mathbf{s}_{\mathrm{cls}}(x)))/\log|\mathcal{C}_{1:t}|$ measures classification uncertainty. Here, $H(\cdot)$ denotes Shannon entropy, which quantifies the uncertainty of a probability distribution, whereas lower entropy indicates a more confident prediction. 
By combining $c(x)$ and $u(x)$, STRIDE increases the contribution of property-based predictions when property confidence is high and the classification pathway is uncertain, using prediction confidence as a proxy for property reliability.
$\operatorname{clip}(\cdot,0,1)$ bounds the fusion weight within $[0,1]$, while $\alpha_{\min}$ and $\beta$ control its minimum value and contribution strength.

In summary, during inference, STRIDE predicts properties according to Eq.~\eqref{eq:property_prediction} and constructs classification evidence from the global property logits according to Eq.~\eqref{eq:property_evidence}. This evidence is then used to enhance visual representations, which are fed into the base classification pathway in Eq.~\eqref{eq:base_classification} to compute class logits. Finally, these logits are adaptively combined with property-to-class predictions according to Eq.~\eqref{eq:property_class_fusion}, yielding the final classification results. The predictions of STRIDE are given by:
\begin{equation}
\hat{\mathbf{p}}_t=f_t^{\mathrm{prop}}(x)
=\left(\arg\max_{v\in\mathcal{P}_k^{(t)}}
[\mathbf{q}_k(x)]_v\right)_{k=1}^{K}.
\label{eq:stride_property_prediction}
\end{equation}
\begin{equation}
\hat{y}_t=f_t^{\mathrm{cls}}(x)
=\arg\max_{c\in\mathcal{C}_{1:t}}
[\mathbf{s}_{\mathrm{final}}(x)]_c.
\label{eq:stride_class_prediction}
\end{equation}

\subsection{Decoupled Class-Property Learning}
Although classification and property prediction share the same visual backbone, learning to predict classes and properties differs from learning to exploit their complementary information. Optimizing the prediction pathways and their interactions simultaneously may destabilize both objectives, as classification must learn from property evidence that is still being established. Therefore, STRIDE adopts a decoupled A-B stage learning strategy for each task.

In stage A, STRIDE establishes the base classification and property prediction paths. The classification path is optimized using cross-entropy with VIB regularization \cite{xie2026area}:
\begin{equation}
\mathcal{L}_{\mathrm{A}}
=
\mathcal{L}_{\mathrm{ce}}^{A}
+
\mathcal{L}_{\mathrm{VIB}},
\label{eq:stage_a_class_loss}
\end{equation}
where $\mathcal{L}_{\mathrm{ce}}^{A}=\mathrm{CE}(\mathbf{s}_{\mathrm{cls}}(x),y)$ supervises the base logits in Eq.~\eqref{eq:base_classification}.
$\mathcal{L}_{\mathrm{A}}$ updates the adapters $\mathcal{A}_v$ and $\mathcal{A}_t$ and the scoring module $\mathbf{s}_t$, $\mathcal{L}_{\mathrm{VIB}}$ regularizes the visual and textual projections within $\mathbf{s}_t$. The CLIP encoders $E_v$ and $E_t$ and historical $\mathbf{s}_j$ ($j<t$) remain frozen. 
Meanwhile, the property predictors $g_k$ and $r_k$ are updated independently of the classification pathway through closed-form solutions based on tasks-accumulated feature statistics:
\begin{equation}
\mathbf{W}_{h,k}^{(t)}
=
\left(\mathbf{G}_{h,k}^{(t)}+\rho_{h,k}\mathbf{I}\right)^{-1}
\mathbf{B}_{h,k}^{(t)},\quad h\in\{g,r\},
\label{eq:stage_a_prop_update}
\end{equation}
where $\mathbf{G}_{h,k}^{(t)}$ accumulates the outer products of feature vectors, and $\mathbf{B}_{h,k}^{(t)}$ accumulates their products with property targets across all observed tasks. For $g_k$, these are derived from global features and property labels; for $r_k$, local features and targets are residualized against the fitted global predictor. 
$\rho_{h,k}$ is the scale-adjusted regularization coefficient selected using a held-out subset of the training data.
$g_k$ is fitted first, followed by $r_k$; they are updated only once per task and remain fixed during training. Parameters of the interaction network $\mathcal{F}_{\mathrm{att}}$ and the property-to-class classifier are not updated in stage A.

In stage B, $\mathcal{A}_v$, $\mathcal{A}_t$, and $\mathbf{s}_t$ are frozen, while $\mathcal{F}_{\mathrm{att}}$ and the property-to-class classifier are optimized. The objective is defined as
\begin{equation}
\mathcal{L}_{\mathrm{B}}
=
\mathcal{L}_{\mathrm{ce}}^{B}
+
\mathcal{L}_{\mathrm{aux}}
+
\mathcal{L}_{\mathrm{pc}},
\label{eq:stage_b_loss}
\end{equation}
where $\mathcal{L}_{\mathrm{ce}}^{B}=\mathrm{CE}(\mathbf{s}_{\mathrm{final}}(x),y)$ supervises the final logits in Eq.~\eqref{eq:property_class_fusion}. The auxiliary loss $\mathcal{L}_{\mathrm{aux}}=\mathrm{CE}(\mathbf{s}_{\mathrm{aux}}(x),y)$ directly supervises the enhanced representation $\mathbf{z}_{\mathrm{enh}}$ through $\mathbf{s}_{\mathrm{aux}}(x)=\mathcal{C}_{\mathrm{aux}}(\mathbf{z}_{\mathrm{enh}})$, where $\mathcal{C}_{\mathrm{aux}}$ is a temporary cosine classifier. This encourages the enhanced representation to maintain class discriminability independently of the final classification logits.
The property-to-class loss $\mathcal{L}_{\mathrm{pc}}=\mathrm{CE}(\mathbf{s}_{\mathrm{prop}}(x),y)$ encourages the property-derived logits to assign higher probability to the ground-truth class, improving the reliability of property-based predictions in final classification.

In summary, stage A learns the base classification and property pathway, while stage B learns to exploit predicted property evidence for representation enhancement and refined class prediction.
Following established practices in PTM-based CL \cite{zhang2023slca,xie2026area}, we estimate class-wise Gaussian distributions from frozen CLIP features of training images and sample labeled features for historical classes. These features are combined with the CLIP features of current training samples to compute $\mathcal{L}_{\mathrm{A}}$ and $\mathcal{L}_{\mathrm{B}}$, preserving discrimination of previously learned classes without storing historical images.

%% file: sec/exp.tex
\section{Experiments}

\begin{table*}[t]
    \centering
    \normalsize
    \setlength{\tabcolsep}{2.7pt}
    \renewcommand{\arraystretch}{1.18}

    \def\stridescore#1#2{\mbox{$#1{\scriptstyle\pm #2}$}}
    \def\stridebest#1#2{\mbox{$\mathbf{#1}{\scriptstyle\pm #2}$}}
    \caption{Final classification accuracy ($A_T$, \%) and property prediction mean average precision ($\mathrm{mAP}^{P}_T$, \%) across four continual learning benchmarks. Best results are highlighted in bold.}
    \label{tab:stride_main}
    \begin{tabularx}{0.9\textwidth}{@{}l*{8}{Y}@{}}
        \toprule
        \multirow{2}{*}{Method}
        & \multicolumn{2}{c}{\shortstack[c]{CIFAR-100\\[-1pt]{\footnotesize 6 tasks}}}
        & \multicolumn{2}{c}{\shortstack[c]{Aircraft\\[-1pt]{\footnotesize 10 tasks}}}
        & \multicolumn{2}{c}{\shortstack[c]{CUB-A\\[-1pt]{\footnotesize 20 tasks}}}
        & \multicolumn{2}{c}{\shortstack[c]{CUB-B\\[-1pt]{\footnotesize 20 tasks}}} \\
        \cmidrule(lr){2-3}
        \cmidrule(lr){4-5}
        \cmidrule(lr){6-7}
        \cmidrule(lr){8-9}
        & $A_T$ & $\mathrm{mAP}^{P}_T$
        & $A_T$ & $\mathrm{mAP}^{P}_T$
        & $A_T$ & $\mathrm{mAP}^{P}_T$
        & $A_T$ & $\mathrm{mAP}^{P}_T$ \\
        \midrule
        TUNA
        & \stridescore{58.01}{0.81} & \stridescore{57.83}{1.02}
        & \stridescore{28.44}{1.23} & \stridescore{24.80}{1.00}
        & \stridescore{56.06}{1.18} & \stridescore{34.98}{1.38}
        & \stridescore{52.54}{1.37} & \stridescore{29.39}{1.35} \\
        
        DualPrompt
        & \stridescore{77.56}{0.56} & \stridescore{85.02}{0.73}
        & \stridescore{55.84}{0.87} & \stridescore{30.11}{0.93}
        & \stridescore{77.06}{0.97} & \stridescore{48.82}{1.12}
        & \stridescore{77.01}{0.95} & \stridescore{42.75}{1.16} \\
        
        RAPF
        & \stridescore{80.88}{0.76} & \stridescore{77.53}{0.93}
        & \stridescore{25.98}{1.45} & \stridescore{30.93}{1.29}
        & \stridescore{48.85}{1.26} & \stridescore{50.17}{1.13}
        & \stridescore{48.85}{1.23} & \stridescore{42.35}{1.15} \\
        
        CLG-CBM
        & \stridescore{80.00}{0.67} & \stridescore{78.23}{0.94}
        & \stridescore{55.30}{1.05} & \stridescore{43.53}{1.02}
        & \stridescore{77.06}{1.13} & \stridescore{35.77}{1.34}
        & \stridescore{77.31}{1.18} & \stridescore{32.05}{1.39} \\

        PROOF
        & \stridescore{78.18}{0.47} & \stridescore{74.46}{0.93}
        & \stridescore{56.74}{0.94} & \stridescore{53.02}{1.08}
        & \stridescore{78.97}{0.83} & \stridescore{37.95}{1.04}
        & \stridescore{78.50}{0.94} & \stridescore{31.33}{1.14} \\
        
        ENGINE
        & \stridescore{79.44}{0.52} & \stridescore{80.55}{0.83}
        & \stridescore{57.97}{0.91} & \stridescore{45.32}{1.13}
        & \stridescore{80.11}{0.85} & \stridescore{45.75}{1.21}
        & \stridescore{80.07}{0.87} & \stridescore{40.61}{1.33} \\

        AREA
        & \stridescore{82.73}{0.49} & \stridescore{56.43}{0.63}
        & \stridescore{61.60}{0.72} & \stridescore{32.73}{0.95}
        & \stridescore{80.79}{0.75} & \stridescore{42.24}{1.09}
        & \stridescore{80.87}{0.83} & \stridescore{36.71}{1.12} \\

        \midrule
        \textbf{STRIDE}
        & \stridebest{83.70}{0.45} & \stridebest{87.31}{0.62}
        & \stridebest{68.41}{0.73} & \stridebest{76.06}{0.92}
        & \stridebest{84.02}{0.77} & \stridebest{54.22}{0.98}
        & \stridebest{84.03}{0.74} & \stridebest{47.11}{1.05} \\

        \bottomrule
    \end{tabularx}
\end{table*}
\begin{figure*}[!t]
  \centering
  \subfloat[Aircraft $A_t$]{
    \includegraphics[height=1.65in,keepaspectratio]{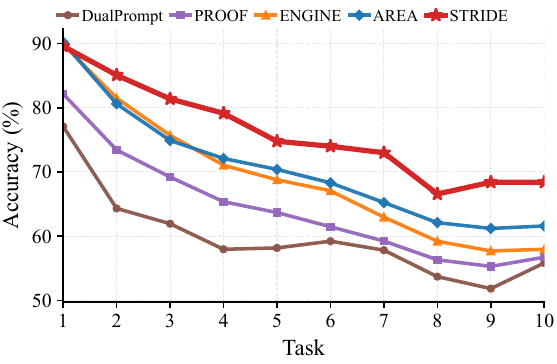}
    }  
  \subfloat[Aircraft $\mathrm{mAP}^{P}_t$]{
    \includegraphics[height=1.65in,keepaspectratio]{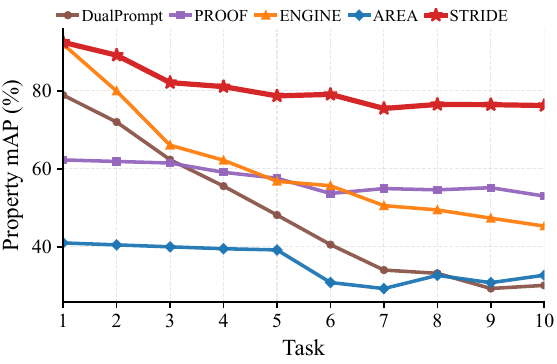}
    }  
    \\
  \subfloat[CUB-A $A_t$]{
    \includegraphics[height=1.65in,keepaspectratio]{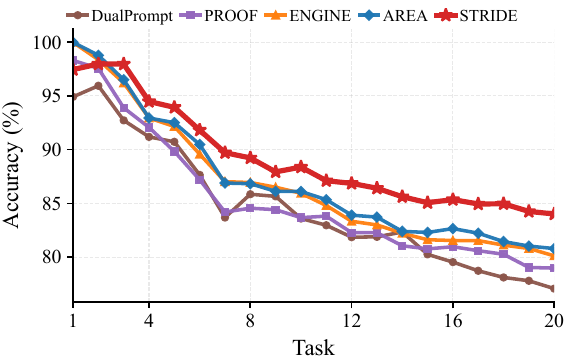}
    }  
  \subfloat[CUB-A $\mathrm{mAP}^{P}_t$]{
    \includegraphics[height=1.65in,keepaspectratio]{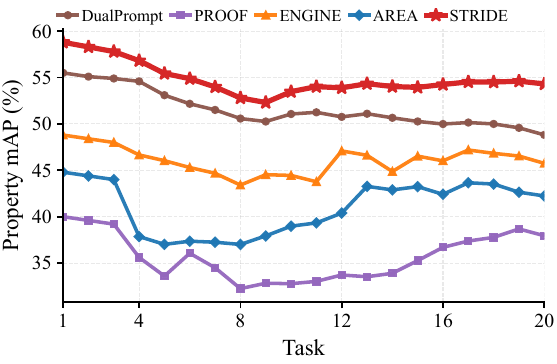}
    }  
  \caption{Task-wise classification accuracy and property mAP of different methods.}
  \label{fig:curve1}
\end{figure*}

\subsection{Implementation Details}
\label{subsec:5.1}
\textbf{Dataset and splits.} We conduct experiments on CIFAR-100 \cite{krizhevsky2009learning}, FGVC-Aircraft \cite{maji2013aircraft}, and CUB-200-2011 \cite{wah2011cub}. We adopt B50 Inc10 for CIFAR-100 (6 tasks), B0 Inc10 for Aircraft (10 tasks), and B10 Inc10 for CUB (20 tasks). For CUB, we construct two variants, CUB-A and CUB-B, using two disjoint property sets with identical class splits. Detailed property definitions are provided in the Supplementary Material.

\noindent\textbf{Baselines.} We compare STRIDE with seven representative CL methods: prompt-based DualPrompt \cite{wang2022dualprompt}, representation adjustment method RAPF \cite{huang2024rapf}, cross-modal projection method PROOF \cite{zhou2025learning}, external knowledge-based ENGINE \cite{zhou2025engine}, concept bottleneck method CLG-CBM \cite{yu2025clgcbm}, adapter-based TUNA \cite{wang2025tuna}, and attribute aggregation method AREA \cite{xie2026area}. All methods are evaluated using identical class splits and property annotations for classification and property prediction.

\noindent\textbf{Programming and hyperparameters.} We implement STRIDE in PyTorch 2.5 based on C3Box \cite{sun2026c3box}, following the original configurations, including hyperparameters and backbone settings of competing methods. To reduce hyperparameter tuning complexity and maintain a consistent optimization setup, all loss weights in stages A and B are fixed to 1 across all experimental settings.

\noindent\textbf{Evaluation metrics.} 
We evaluate both classification and property prediction performance after the final task. The final classification accuracy $A_T$ is measured over the joint test set of all seen classes. For property prediction, we report the mean average precision $\mathrm{mAP}^{P}_T$ across all property dimensions. Specifically, average precision (AP) is computed for each property value $v\in\mathcal{P}_k^{(T)}$ by ranking test samples according to the predicted probability $[q_k(x)]_v$ in a one-vs-rest manner. The property mAP is defined as:
\begin{equation}
\mathrm{mAP}^{P}_T =
\frac{
\sum_{k=1}^{K}
\sum_{v\in\mathcal{P}_k^{+}}
\mathrm{AP}_T(k,v)
}{
\sum_{k=1}^{K}|\mathcal{P}_k^{+}|
},
\end{equation}
where $\mathcal{P}_k^{+}$ denotes the property values with positive test instances, and $\mathrm{AP}_T(k,v)$ is the AP for value $v$ in dimension $k$. 
For fair comparison, we equip baseline methods with fully connected property prediction heads trained using cross-entropy loss, following conventional supervised attribute prediction \cite{wu2016visualconcepts}. All property-supervised variants use identical property annotations and supervision. For each baseline and random seed, we report the higher classification accuracy between its original and property-supervised variants to avoid penalizing baselines for potential degradation caused by additional property supervision. Property mAP is always evaluated using the property-supervised variant. This protocol ensures that classification comparisons are made against the stronger baseline variant, rather than benefiting from any performance degradation introduced by additional property supervision.

\subsection{Benchmark Comparison}
\label{subsec:benchmark_comparison}

This subsection evaluates STRIDE against representative CL methods across four benchmark settings.
Table~\ref{tab:stride_main} reports the mean and standard deviation of final classification accuracy $A_T$ and property prediction $\mathrm{mAP}^{P}_T$ across four benchmark settings, with all methods evaluated using the same five fixed random seeds.
STRIDE consistently outperforms all competing methods in both classification accuracy and property prediction across the four benchmarks. Notably, existing methods exhibit varying strengths in classification and property prediction, whereas STRIDE consistently achieves superior performance on both objectives. 
Fig.~\ref{fig:curve1} illustrates the task-wise trajectories of classification accuracy and property mAP on Aircraft and CUB-A. STRIDE achieves superior overall performance as tasks accumulate, ultimately attaining the highest classification accuracy and property mAP on both benchmarks. These results demonstrate the effectiveness of STRIDE in jointly predicting classes and properties while maintaining strong classification performance throughout continual learning.

\subsection{Ablation Study}

\begin{table}[t]
    \centering
    \footnotesize
    \setlength{\tabcolsep}{2pt}
    \renewcommand{\arraystretch}{1.2}

    \newcommand{\abval}[2]{%
        $#1{\scriptscriptstyle\pm #2}$}
    \caption{Ablation study of STRIDE.}
    \label{tab:stride_ablation}
    \begin{tabularx}{\columnwidth}{
        @{}
        >{\raggedright\arraybackslash}p{0.31\columnwidth}
        *{4}{>{\centering\arraybackslash}X}
        @{}
    }
        \toprule
        \multirow{2}{*}{Variant}
        & \multicolumn{2}{c}{Aircraft}
        & \multicolumn{2}{c}{CUB-A} \\
        \cmidrule(lr){2-3}
        \cmidrule(lr){4-5}
        & $A_T$ & $\mathrm{mAP}^{P}_T$
        & $A_T$ & $\mathrm{mAP}^{P}_T$ \\
        \midrule

        Joint Stage A-B
        & \abval{60.67}{0.88}
        & \abval{76.06}{0.92}
        & \abval{81.34}{0.89}
        & \abval{54.22}{0.98} \\

        w/o $\mathcal{L}_{\mathrm{aux}}$
        & \abval{66.17}{0.84}
        & \abval{76.06}{0.92}
        & \abval{82.71}{0.82}
        & \abval{54.22}{0.98} \\

        w/o Prop
        & \abval{62.05}{0.78}
        & -
        & \abval{80.88}{0.76}
        & - \\
        
        w/o GLP
        & \abval{67.32}{0.83}
        & \abval{55.22}{0.95}
        & \abval{82.88}{0.88}
        & \abval{37.51}{1.06} \\

        w/o RE
        & \abval{67.53}{0.87}
        & \abval{76.06}{0.95}
        & \abval{83.15}{0.84}
        & \abval{54.22}{1.02} \\

        w/o P2C
        & \abval{63.71}{0.85}
        & \abval{76.06}{0.92}
        & \abval{81.63}{0.78}
        & \abval{54.22}{0.98} \\

        \midrule

        \textbf{STRIDE}
        & \textbf{\abval{68.41}{0.73}}
        & \textbf{\abval{76.06}{0.92}}
        & \textbf{\abval{84.02}{0.77}}
        & \textbf{\abval{54.22}{0.98}} \\

        \bottomrule
    \end{tabularx}
\end{table}

We conduct ablation studies on Aircraft and CUB-A to evaluate the effectiveness of STRIDE's key components. Table~\ref{tab:stride_ablation} summarizes the results.
First, we investigate the decoupled training strategy. Jointly optimizing the objectives of stages A and B (Joint Stage A-B) leads to substantial classification accuracy drops of 7.74 and 2.68 percentage points on Aircraft and CUB-A, respectively. This suggests that simultaneously updating parameters originally optimized in separate stages introduces unstable interactions between base pathway learning and property-guided classification refinement, leading to degraded classification performance. Removing the auxiliary classification loss ($\mathcal{L}_{\mathrm{aux}}$) reduces classification accuracy, highlighting the importance of directly supervising the discriminative capability of enhanced visual representations. Unlike disabling representation enhancement entirely, removing $\mathcal{L}{\mathrm{aux}}$ retains the enhancement module but weakens its direct discriminative supervision, potentially introducing less effective feature modifications into the classification pathway.
We then examine the contribution of property prediction. Removing property prediction entirely (w/o Prop) decreases classification accuracy by 6.36 and 3.14 percentage points on Aircraft and CUB-A, respectively, highlighting the importance of incorporating properties into continual learning. Replacing the proposed global-local analytical property classifier with a visual-feature-based linear classifier (w/o GLP) substantially reduces property mAP and classification accuracy, demonstrating the effectiveness of our analytical property prediction mechanism and its contribution to classification.
Finally, we investigate the two components of property-guided classification enhancement. Removing representation enhancement (w/o RE) or the property-to-class classifier (w/o P2C) reduces classification accuracy. These results demonstrate that property evidence contributes to classification through both representation enhancement and complementary class prediction. 

\subsection{Analysis of Property-Guided Classification}
\label{subsec:paired_prediction}
\begin{figure}[t]
    \centering
    \includegraphics[width=0.95\columnwidth]{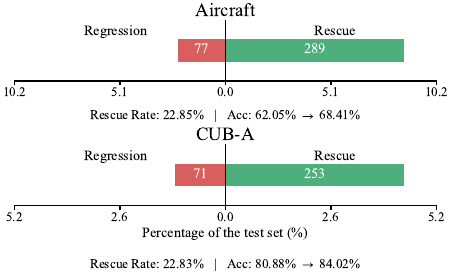}
    \caption{Sample-level classification changes with property learning. The horizontal axis indicates the percentages of rescue and regression cases relative to the total number of test samples.}
    \label{fig:paired_switches}
\end{figure}

To further investigate how explicit property learning improves classification, we conduct a sample-level paired analysis between STRIDE and its counterpart without property learning (w/o Prop). We compare their predictions on identical test images and examine both the corrected classification errors and the new errors introduced by property learning. Specifically, a rescue denotes a sample misclassified by w/o Prop but correctly classified by STRIDE, whereas a regression denotes a sample correctly classified by w/o Prop but misclassified by STRIDE. The rescue rate measures the proportion of errors made by w/o Prop that STRIDE corrects. 
As shown in Fig.~\ref{fig:paired_switches}, on Aircraft, incorporating property learning achieves a rescue rate of 22.85\%, while introducing only 77 regressions. This improves classification accuracy from 62.05\% to 68.41\%. Similarly, on CUB-A, property learning achieves a rescue rate of 22.83\%, with only 71 regressions, improving accuracy from 80.88\% to 84.02\%. On both benchmarks, the number of rescued predictions substantially exceeds that of regressions, indicating that explicit property learning corrects a considerable portion of previously misclassified samples while introducing relatively few new errors. These results provide sample-level evidence that property learning effectively improves continual classification.

\begin{figure}[!t]
  \centering
  \subfloat[Pred: A340-300]{
    \includegraphics[height=1.1in,keepaspectratio]{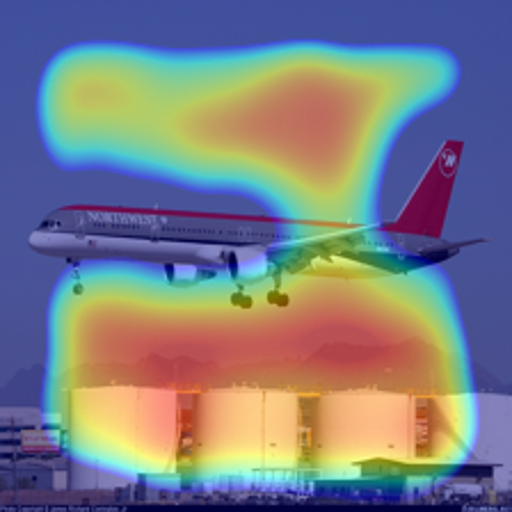}
    }  
  \subfloat[Pred: 757-300]{
    \includegraphics[height=1.1in,keepaspectratio]{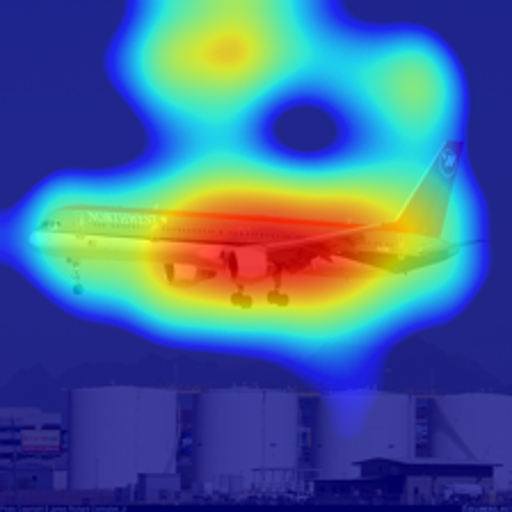}
    }  
    \\
  \subfloat[Pred: Cowbird]{
    \includegraphics[height=1.1in,keepaspectratio]{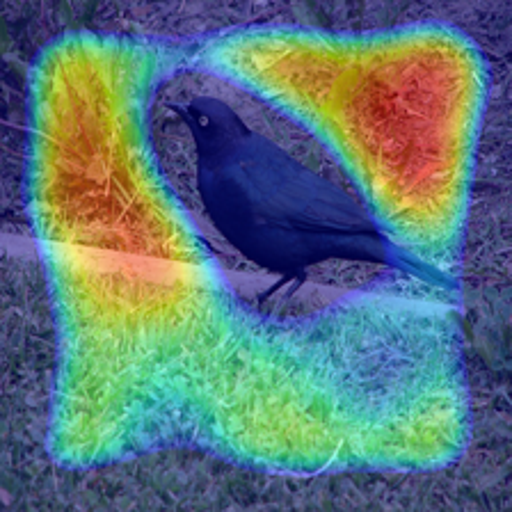}
    }  
  \subfloat[Pred: Blackbird]{
    \includegraphics[height=1.1in,keepaspectratio]{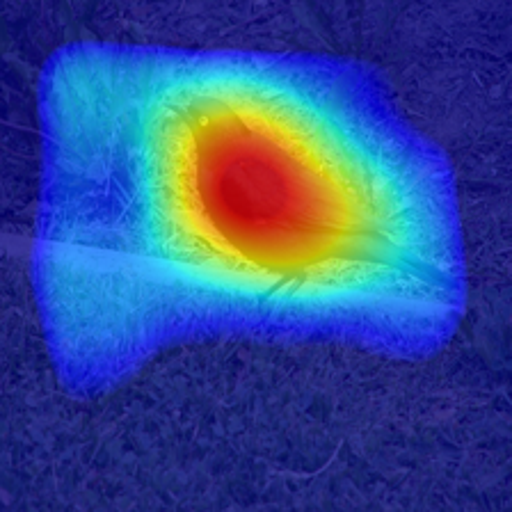}
    }  
    \\
  \subfloat[Pred: Albatross]{
    \includegraphics[height=1.1in,keepaspectratio]{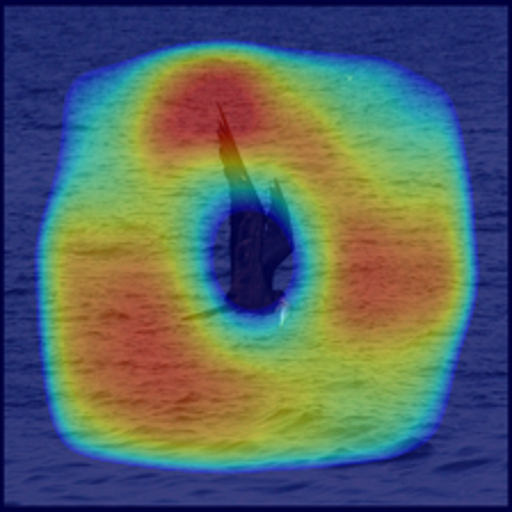}
    }  
  \subfloat[Pred: Frigatebird]{
    \includegraphics[height=1.1in,keepaspectratio]{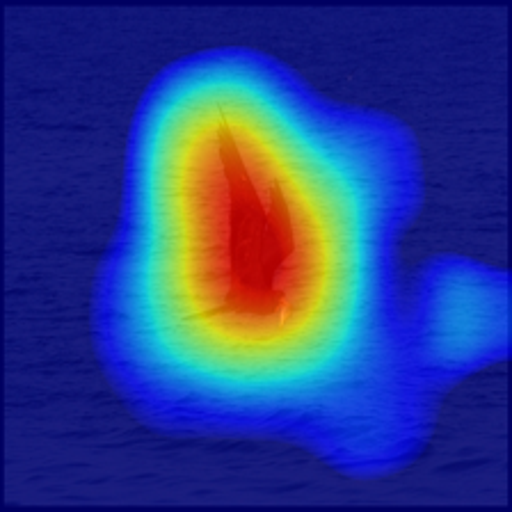}
    }  
  \caption{Class activation maps without (left) and with (right) property-guided learning.}
  \label{fig:vis}
\end{figure}

To visualize the effect of property-guided learning, Fig.~\ref{fig:vis} presents class activation maps for representative samples. The results suggest that property prediction helps the model better focus on discriminative regions of target objects rather than irrelevant backgrounds, potentially leading to more informative visual representations for classification. Furthermore, predicted properties can provide classification evidence through the property-to-class pathway, further enhancing the model's discriminative capability. For example, w/o Prop misclassifies a Brewer Blackbird as a Shiny Cowbird while exhibiting strong responses to the surrounding background, whereas STRIDE shows more concentrated activations on the bird's body and plumage and correctly identifies its class.
Further experiments in the supplementary material demonstrate that STRIDE maintains robust classification performance even when property predictions are inaccurate, highlighting its ability to mitigate the influence of unreliable property evidence.

%% file: sec/conclu.tex
\section{Conclusion}
This paper introduces STRIDE, a framework that jointly performs classification and property prediction for CLIP-based continual learning. STRIDE establishes a classification pathway alongside global-local property predictors, which are analytically updated using accumulated statistics to preserve property knowledge across tasks. It further transforms predicted properties into discriminative evidence, enhancing visual representations and providing complementary class predictions through reliability-aware interaction. A two-stage learning strategy separates the establishment of class and property predictors from property-guided classification enhancement, reducing interference between their learning objectives. Extensive experiments across multiple continual learning benchmarks demonstrate that STRIDE improves classification performance over strong baselines while maintaining competitive property prediction. Future work will explore more robust property estimation and adaptive interaction strategies to further improve both prediction objectives.

%% file: main.bib
@String(CVPR= {IEEE Conf. Comput. Vis. Pattern Recog.})

@String(ICCV= {Int. Conf. Comput. Vis.})

@String(ECCV= {Eur. Conf. Comput. Vis.})

@String(AAAI = {AAAI})

@String(CVPR  = {CVPR})

@String(ICCV  = {ICCV})

@String(ECCV  = {ECCV})

@inproceedings{chen2024promptfusion,
  title        = {{PromptFusion}: Decoupling Stability and Plasticity for Continual Learning},
  author       = {Chen, Haoran and Wu, Zuxuan and Han, Xintong and Jia, Menglin and Jiang, Yu-Gang},
  booktitle    = {European Conference on Computer Vision (ECCV)},
  year         = {2024},
  url          = {https://www.ecva.net/papers/eccv_2024/papers_ECCV/html/11826_ECCV_2024_paper.php}
}

@inproceedings{zheng2025tagfex,
  title        = {Task-Agnostic Guided Feature Expansion for Class-Incremental Learning},
  author       = {Zheng, Bowen and Zhou, Da-Wei and Ye, Han-Jia and Zhan, De-Chuan},
  booktitle    = {Proceedings of the IEEE/CVF Conference on Computer Vision and Pattern Recognition (CVPR)},
  pages        = {10099--10109},
  year         = {2025},
  url          = {https://openaccess.thecvf.com/content/CVPR2025/html/Zheng_Task-Agnostic_Guided_Feature_Expansion_for_Class-Incremental_Learning_CVPR_2025_paper.html}
}

@inproceedings{gao2024cprompt,
  title        = {Consistent Prompting for Rehearsal-Free Continual Learning},
  author       = {Gao, Zhanxin and Cen, Jun and Chang, Xiaobin},
  booktitle    = {Proceedings of the IEEE/CVF Conference on Computer Vision and Pattern Recognition (CVPR)},
  pages        = {28463--28473},
  year         = {2024},
  url          = {https://openaccess.thecvf.com/content/CVPR2024/html/Gao_Consistent_Prompting_for_Rehearsal-Free_Continual_Learning_CVPR_2024_paper.html}
}

@inproceedings{wang2025sema,
  title        = {Self-Expansion of Pre-trained Models with Mixture of Adapters for Continual Learning},
  author       = {Wang, Huiyi and Lu, Haodong and Yao, Lina and Gong, Dong},
  booktitle    = {Proceedings of the IEEE/CVF Conference on Computer Vision and Pattern Recognition (CVPR)},
  pages        = {10087--10098},
  year         = {2025},
  url          = {https://openaccess.thecvf.com/content/CVPR2025/html/Wang_Self-Expansion_of_Pre-trained_Models_with_Mixture_of_Adapters_for_Continual_CVPR_2025_paper.html}
}

@inproceedings{gao2025moal,
  title        = {Knowledge Memorization and Rumination for Pre-trained Model-based Class-Incremental Learning},
  author       = {Gao, Zijian and Jia, Wangwang and Zhang, Xingxing and Zhou, Dulan and Xu, Kele and Feng, Dawei and Dou, Yong and Mao, Xinjun and Wang, Huaimin},
  booktitle    = {Proceedings of the IEEE/CVF Conference on Computer Vision and Pattern Recognition (CVPR)},
  pages        = {20523--20533},
  year         = {2025},
  url          = {https://openaccess.thecvf.com/content/CVPR2025/html/Gao_Knowledge_Memorization_and_Rumination_for_Pre-trained_Model-based_Class-Incremental_Learning_CVPR_2025_paper.html}
}

@inproceedings{wang2025tuna,
  title        = {Integrating Task-Specific and Universal Adapters for Pre-Trained Model-based Class-Incremental Learning},
  author       = {Wang, Yan and Zhou, Da-Wei and Ye, Han-Jia},
  booktitle    = {Proceedings of the IEEE/CVF International Conference on Computer Vision (ICCV)},
  pages        = {806--816},
  year         = {2025},
  doi          = {10.1109/ICCV51701.2025.00083},
  url          = {https://openaccess.thecvf.com/content/ICCV2025/html/Wang_Integrating_Task-Specific_and_Universal_Adapters_for_Pre-Trained_Model-based_Class-Incremental_Learning_ICCV_2025_paper.html}
}

@inproceedings{radford2021clip,
  title        = {Learning Transferable Visual Models From Natural Language Supervision},
  author       = {Radford, Alec and Kim, Jong Wook and Hallacy, Chris and Ramesh, Aditya and Goh, Gabriel and Agarwal, Sandhini and Sastry, Girish and Askell, Amanda and Mishkin, Pamela and Clark, Jack and Krueger, Gretchen and Sutskever, Ilya},
  booktitle    = {Proceedings of the 38th International Conference on Machine Learning (ICML)},
  series       = {Proceedings of Machine Learning Research},
  volume       = {139},
  pages        = {8748--8763},
  year         = {2021},
  url          = {https://proceedings.mlr.press/v139/radford21a.html}
}

@inproceedings{wang2023attriclip,
  title        = {{AttriCLIP}: A Non-Incremental Learner for Incremental Knowledge Learning},
  author       = {Wang, Runqi and Duan, Xiaoyue and Kang, Guoliang and Liu, Jianzhuang and Lin, Shaohui and Xu, Songcen and L{\"u}, Jinhu and Zhang, Baochang},
  booktitle    = {Proceedings of the IEEE/CVF Conference on Computer Vision and Pattern Recognition (CVPR)},
  pages        = {3654--3663},
  year         = {2023},
  url          = {https://openaccess.thecvf.com/content/CVPR2023/html/Wang_AttriCLIP_A_Non-Incremental_Learner_for_Incremental_Knowledge_Learning_CVPR_2023_paper.html}
}

@inproceedings{huang2024rapf,
  title        = {Class-Incremental Learning with {CLIP}: Adaptive Representation Adjustment and Parameter Fusion},
  author       = {Huang, Linlan and Cao, Xusheng and Lu, Haori and Liu, Xialei},
  booktitle    = {European Conference on Computer Vision (ECCV)},
  year         = {2024},
  url          = {https://www.ecva.net/papers/eccv_2024/papers_ECCV/html/7092_ECCV_2024_paper.php}
}

@inproceedings{yu2024moeadapters,
  title        = {Boosting Continual Learning of Vision-Language Models via Mixture-of-Experts Adapters},
  author       = {Yu, Jiazuo and Zhuge, Yunzhi and Zhang, Lu and Hu, Ping and Wang, Dong and Lu, Huchuan and He, You},
  booktitle    = {Proceedings of the IEEE/CVF Conference on Computer Vision and Pattern Recognition (CVPR)},
  pages        = {23219--23230},
  year         = {2024},
  url          = {https://openaccess.thecvf.com/content/CVPR2024/html/Yu_Boosting_Continual_Learning_of_Vision-Language_Models_via_Mixture-of-Experts_Adapters_CVPR_2024_paper.html}
}

@inproceedings{luo2025lada,
  title        = {{LADA}: Scalable Label-Specific {CLIP} Adapter for Continual Learning},
  author       = {Luo, Mao-Lin and Zhou, Zi-Hao and Wei, Tong and Zhang, Min-Ling},
  booktitle    = {Proceedings of the 42nd International Conference on Machine Learning (ICML)},
  series       = {Proceedings of Machine Learning Research},
  volume       = {267},
  pages        = {41604--41619},
  year         = {2025},
  url          = {https://proceedings.mlr.press/v267/luo25w.html}
}

@inproceedings{huang2025mgclip,
  title        = {Mind the Gap: Preserving and Compensating for the Modality Gap in {CLIP}-Based Continual Learning},
  author       = {Huang, Linlan and Cao, Xusheng and Lu, Haori and Meng, Yifan and Yang, Fei and Liu, Xialei},
  booktitle    = {Proceedings of the IEEE/CVF International Conference on Computer Vision (ICCV)},
  pages        = {3777--3786},
  year         = {2025},
  url          = {https://openaccess.thecvf.com/content/ICCV2025/html/Huang_Mind_the_Gap_Preserving_and_Compensating_for_the_Modality_Gap_ICCV_2025_paper.html}
}

@inproceedings{lin2023pcr,
  title = {{PCR}: Proxy-Based Contrastive Replay for Online Class-Incremental Continual Learning},
  author = {Lin, Huiwei and Zhang, Baoquan and Feng, Shanshan and Li, Xutao and Ye, Yunming},
  booktitle = {Proceedings of the IEEE/CVF Conference on Computer Vision and Pattern Recognition (CVPR)},
  pages = {24246--24255},
  year = {2023}
}

@inproceedings{yoo2024lpr,
  title = {Layerwise Proximal Replay: A Proximal Point Method for Online Continual Learning},
  author = {Yoo, Jinsoo and Liu, Yunpeng and Wood, Frank and Pleiss, Geoff},
  booktitle = {Proceedings of the 41st International Conference on Machine Learning},
  series = {Proceedings of Machine Learning Research},
  volume = {235},
  pages = {57199--57216},
  publisher = {PMLR},
  year = {2024},
  url = {https://proceedings.mlr.press/v235/yoo24a.html}
}

@inproceedings{asadi2023psrd,
  title = {Prototype-Sample Relation Distillation: Towards Replay-Free Continual Learning},
  author = {Asadi, Nader and Davari, Mohammadreza and Mudur, Sudhir and Aljundi, Rahaf and Belilovsky, Eugene},
  booktitle = {Proceedings of the 40th International Conference on Machine Learning},
  series = {Proceedings of Machine Learning Research},
  volume = {202},
  pages = {1093--1106},
  publisher = {PMLR},
  year = {2023},
  url = {https://proceedings.mlr.press/v202/asadi23a.html}
}

@inproceedings{wen2024mtd,
  title = {Class Incremental Learning with Multi-Teacher Distillation},
  author = {Wen, Haitao and Pan, Lili and Dai, Yu and Qiu, Heqian and Wang, Lanxiao and Wu, Qingbo and Li, Hongliang},
  booktitle = {Proceedings of the IEEE/CVF Conference on Computer Vision and Pattern Recognition (CVPR)},
  pages = {28443--28452},
  year = {2024}
}

@inproceedings{zhou2024ease,
  title = {Expandable Subspace Ensemble for Pre-Trained Model-Based Class-Incremental Learning},
  author = {Zhou, Da-Wei and Sun, Hai-Long and Ye, Han-Jia and Zhan, De-Chuan},
  booktitle = {Proceedings of the IEEE/CVF Conference on Computer Vision and Pattern Recognition (CVPR)},
  pages = {23554--23564},
  year = {2024}
}

@inproceedings{li2025dia,
  title = {Dynamic Integration of Task-Specific Adapters for Class Incremental Learning},
  author = {Li, Jiashuo and Wang, Shaokun and Qian, Bo and He, Yuhang and Wei, Xing and Wang, Qiang and Gong, Yihong},
  booktitle = {Proceedings of the IEEE/CVF Conference on Computer Vision and Pattern Recognition (CVPR)},
  pages = {30545--30555},
  year = {2025}
}

@inproceedings{smith2023coda,
  title = {{CODA-Prompt}: COntinual Decomposed Attention-Based Prompting for Rehearsal-Free Continual Learning},
  author = {Smith, James Seale and Karlinsky, Leonid and Gutta, Vyshnavi and Cascante-Bonilla, Paola and Kim, Donghyun and Arbelle, Assaf and Panda, Rameswar and Feris, Rogerio and Kira, Zsolt},
  booktitle = {Proceedings of the IEEE/CVF Conference on Computer Vision and Pattern Recognition (CVPR)},
  pages = {11909--11919},
  year = {2023}
}

@inproceedings{zhang2023slca,
  title = {{SLCA}: Slow Learner with Classifier Alignment for Continual Learning on a Pre-trained Model},
  author = {Zhang, Gengwei and Wang, Liyuan and Kang, Guoliang and Chen, Ling and Wei, Yunchao},
  booktitle = {Proceedings of the IEEE/CVF International Conference on Computer Vision (ICCV)},
  pages = {19148--19158},
  year = {2023}
}

@inproceedings{mcdonnell2023ranpac,
  title = {{RanPAC}: Random Projections and Pre-trained Models for Continual Learning},
  author = {McDonnell, Mark D. and Gong, Dong and Parvaneh, Amin and Abbasnejad, Ehsan and van den Hengel, Anton},
  booktitle = {Advances in Neural Information Processing Systems},
  volume = {36},
  year = {2023},
  url = {https://proceedings.neurips.cc/paper_files/paper/2023/hash/2793dc35e14003dd367684d93d236847-Abstract-Conference.html}
}

@inproceedings{he2025cllora,
  title = {{CL-LoRA}: Continual Low-Rank Adaptation for Rehearsal-Free Class-Incremental Learning},
  author = {He, Jiangpeng and Duan, Zhihao and Zhu, Fengqing},
  booktitle = {Proceedings of the IEEE/CVF Conference on Computer Vision and Pattern Recognition (CVPR)},
  pages = {30534--30544},
  year = {2025},
  doi = {10.1109/CVPR52734.2025.02843}
}

@inproceedings{zhao2026rsiat,
  title = {Representation-Steered Incremental Adapter-Tuning for Class-Incremental Learning with Pre-Trained Models},
  author = {Zhao, Jiarui and Huang, Libo and Li, Xiangqi and An, Zhulin and Yang, Chuanguang and Wang, Yu and Diao, Boyu and Xu, Yongjun},
  booktitle = {Proceedings of the IEEE/CVF Conference on Computer Vision and Pattern Recognition (CVPR)},
  pages = {18010--18020},
  year = {2026}
}

@misc{maji2013aircraft,
  title = {Fine-Grained Visual Classification of Aircraft},
  author = {Maji, Subhransu and Kannala, Juho and Rahtu, Esa and Blaschko, Matthew B. and Vedaldi, Andrea},
  year = {2013},
  howpublished = {arXiv preprint arXiv:1306.5151},
  eprint = {1306.5151},
  archivePrefix = {arXiv},
  url = {https://arxiv.org/abs/1306.5151}
}

@techreport{wah2011cub,
  title = {The {Caltech-UCSD Birds-200-2011} Dataset},
  author = {Wah, Catherine and Branson, Steve and Welinder, Peter and Perona, Pietro and Belongie, Serge},
  institution = {California Institute of Technology},
  type = {Computation and Neural Systems Technical Report},
  number = {CNS-TR-2011-001},
  year = {2011},
  url = {https://authors.library.caltech.edu/records/cvm3y-5hh21}
}

@inproceedings{zhou2025engine,
  title = {External Knowledge Injection for {CLIP}-Based Class-Incremental Learning},
  author = {Zhou, Da-Wei and Li, Kai-Wen and Ning, Jingyi and Ye, Han-Jia and Zhang, Lijun and Zhan, De-Chuan},
  booktitle = {Proceedings of the IEEE/CVF International Conference on Computer Vision (ICCV)},
  pages = {3314--3325},
  year = {2025},
  url = {https://openaccess.thecvf.com/content/ICCV2025/html/Zhou_External_Knowledge_Injection_for_CLIP-Based_Class-Incremental_Learning_ICCV_2025_paper.html}
}

@inproceedings{xie2026area,
  title = {{AREA}: Attribute Extraction and Aggregation for {CLIP}-Based Class-Incremental Learning},
  author = {Xie, Zhen-Hao and Shi, Yu-Cheng and Zhou, Da-Wei},
  booktitle = {Proceedings of the 43rd International Conference on Machine Learning},
  series = {Proceedings of Machine Learning Research},
  volume = {306},
  pages = {139637--139658},
  publisher = {PMLR},
  year = {2026},
  url = {https://proceedings.mlr.press/v306/xie26u.html}
}

@article{zhou2025learning,
  title = {Learning without Forgetting for Vision-Language Models},
  author = {Zhou, Da-Wei and Zhang, Yuanhan and Wang, Yan and Ning, Jingyi and Ye, Han-Jia and Zhan, De-Chuan and Liu, Ziwei},
  journal = {IEEE Transactions on Pattern Analysis and Machine Intelligence},
  volume = {47},
  number = {6},
  pages = {4489--4504},
  year = {2025},
  doi = {10.1109/TPAMI.2025.3540889},
  url = {https://doi.org/10.1109/TPAMI.2025.3540889}
}

@inproceedings{wu2025gift,
  title = {Synthetic Data is an Elegant {GIFT} for Continual Vision-Language Models},
  author = {Wu, Bin and Shi, Wuxuan and Wang, Jinqiao and Ye, Mang},
  booktitle = {Proceedings of the IEEE/CVF Conference on Computer Vision and Pattern Recognition (CVPR)},
  pages = {2813--2823},
  year = {2025},
  url = {https://openaccess.thecvf.com/content/CVPR2025/html/Wu_Synthetic_Data_is_an_Elegant_GIFT_for_Continual_Vision-Language_Models_CVPR_2025_paper.html}
}

@inproceedings{yu2025clgcbm,
  title = {Language Guided Concept Bottleneck Models for Interpretable Continual Learning},
  author = {Yu, Lu and Han, Haoyu and Tao, Zhe and Yao, Hantao and Xu, Changsheng},
  booktitle = {Proceedings of the IEEE/CVF Conference on Computer Vision and Pattern Recognition (CVPR)},
  pages = {14976--14986},
  year = {2025},
  url = {https://openaccess.thecvf.com/content/CVPR2025/html/Yu_Language_Guided_Concept_Bottleneck_Models_for_Interpretable_Continual_Learning_CVPR_2025_paper.html}
}

@article{li2026bofa,
  title = {{BOFA}: Bridge-Layer Orthogonal Low-Rank Fusion for {CLIP}-Based Class-Incremental Learning},
  author = {Li, Lan and Hu, Tao and Zhou, Da-Wei and Yang, Jia-Qi and Ye, Han-Jia and Zhan, De-Chuan},
  journal = {Proceedings of the AAAI Conference on Artificial Intelligence},
  volume = {40},
  number = {27},
  pages = {22967--22975},
  year = {2026},
  doi = {10.1609/aaai.v40i27.39461},
  url = {https://ojs.aaai.org/index.php/AAAI/article/view/39461}
}

@inproceedings{luo2023ciec,
  title = {Class-Incremental Exemplar Compression for Class-Incremental Learning},
  author = {Luo, Zilin and Liu, Yaoyao and Schiele, Bernt and Sun, Qianru},
  booktitle = {Proceedings of the IEEE/CVF Conference on Computer Vision and Pattern Recognition (CVPR)},
  pages = {11371--11380},
  year = {2023},
  url = {https://openaccess.thecvf.com/content/CVPR2023/html/Luo_Class-Incremental_Exemplar_Compression_for_Class-Incremental_Learning_CVPR_2023_paper.html}
}

@inproceedings{li2024fcs,
  title = {{FCS}: Feature Calibration and Separation for Non-Exemplar Class Incremental Learning},
  author = {Li, Qiwei and Peng, Yuxin and Zhou, Jiahuan},
  booktitle = {Proceedings of the IEEE/CVF Conference on Computer Vision and Pattern Recognition (CVPR)},
  pages = {28495--28504},
  year = {2024},
  url = {https://openaccess.thecvf.com/content/CVPR2024/html/Li_FCS_Feature_Calibration_and_Separation_for_Non-Exemplar_Class_Incremental_Learning_CVPR_2024_paper.html}
}

@inproceedings{liu2024tass,
  title = {Task-Adaptive Saliency Guidance for Exemplar-free Class Incremental Learning},
  author = {Liu, Xialei and Zhai, Jiang-Tian and Bagdanov, Andrew D. and Li, Ke and Cheng, Ming-Ming},
  booktitle = {Proceedings of the IEEE/CVF Conference on Computer Vision and Pattern Recognition (CVPR)},
  pages = {23954--23963},
  year = {2024},
  url = {https://openaccess.thecvf.com/content/CVPR2024/html/Liu_Task-Adaptive_Saliency_Guidance_for_Exemplar-free_Class_Incremental_Learning_CVPR_2024_paper.html}
}

@inproceedings{hu2023dne,
  title = {Dense Network Expansion for Class Incremental Learning},
  author = {Hu, Zhiyuan and Li, Yunsheng and Lyu, Jiancheng and Gao, Dashan and Vasconcelos, Nuno},
  booktitle = {Proceedings of the IEEE/CVF Conference on Computer Vision and Pattern Recognition (CVPR)},
  pages = {11858--11867},
  year = {2023},
  url = {https://openaccess.thecvf.com/content/CVPR2023/html/Hu_Dense_Network_Expansion_for_Class_Incremental_Learning_CVPR_2023_paper.html}
}

@inproceedings{wang2022dualprompt,
  title = {{DualPrompt}: Complementary Prompting for Rehearsal-Free Continual Learning},
  author = {Wang, Zifeng and Zhang, Zizhao and Ebrahimi, Sayna and Sun, Ruoxi and Zhang, Han and Lee, Chen-Yu and Ren, Xiaoqi and Su, Guolong and Perot, Vincent and Dy, Jennifer and Pfister, Tomas},
  booktitle = {Computer Vision -- ECCV 2022},
  series = {Lecture Notes in Computer Science},
  volume = {13686},
  pages = {631--648},
  publisher = {Springer},
  year = {2022},
  doi = {10.1007/978-3-031-19809-0_36},
  url = {https://doi.org/10.1007/978-3-031-19809-0_36}
}

@inproceedings{ijcai2024p0924,
  title = {Continual Learning with Pre-Trained Models: A Survey},
  author = {Zhou, Da-Wei and Sun, Hai-Long and Ning, Jingyi and Ye, Han-Jia and Zhan, De-Chuan},
  booktitle = {Proceedings of the Thirty-Third International Joint Conference on Artificial Intelligence},
  pages = {8363--8371},
  publisher = {International Joint Conferences on Artificial Intelligence Organization},
  year = {2024},
  doi = {10.24963/ijcai.2024/924},
  url = {https://doi.org/10.24963/ijcai.2024/924}
}

@article{zhou2024class,
  title = {Class-Incremental Learning: A Survey},
  author = {Zhou, Da-Wei and Wang, Qi-Wei and Qi, Zhi-Hong and Ye, Han-Jia and Zhan, De-Chuan and Liu, Ziwei},
  journal = {IEEE Transactions on Pattern Analysis and Machine Intelligence},
  volume = {46},
  number = {12},
  pages = {9851--9873},
  year = {2024},
  doi = {10.1109/TPAMI.2024.3429383},
  url = {https://doi.org/10.1109/TPAMI.2024.3429383}
}

@inproceedings{zhang2025adaptive,
  title = {Adaptive Parameter Selection for Tuning Vision-Language Models},
  author = {Zhang, Yi and Deng, Yi-Xuan and Guo, Meng-Hao and Hu, Shi-Min},
  booktitle = {Proceedings of the IEEE/CVF Conference on Computer Vision and Pattern Recognition (CVPR)},
  pages = {4280--4290},
  year = {2025},
  doi = {10.1109/CVPR52734.2025.00404},
  url = {https://openaccess.thecvf.com/content/CVPR2025/html/Zhang_Adaptive_Parameter_Selection_for_Tuning_Vision-Language_Models_CVPR_2025_paper.html}
}

@inproceedings{wang2024baseline,
  title={A Hard-to-Beat Baseline for {Training-free CLIP-based Adaptation}},
  author={Wang, Zhengbo and Liang, Jian and Sheng, Lijun and He, Ran and Wang, Zilei and Tan, Tieniu},
  booktitle={International Conference on Learning Representations},
  year={2024}
}

@article{hoerl1970ridge,
  title={Ridge Regression: Biased Estimation for Nonorthogonal Problems},
  author={Hoerl, Arthur E. and Kennard, Robert W.},
  journal={Technometrics},
  volume={12},
  number={1},
  pages={55--67},
  year={1970},
  doi={10.1080/00401706.1970.10488634}
}

@techreport{krizhevsky2009learning,
  title       = {Learning Multiple Layers of Features from Tiny Images},
  author      = {Krizhevsky, Alex},
  institution = {University of Toronto},
  year        = {2009},
  url         = {https://www.cs.toronto.edu/~kriz/learning-features-2009-TR.pdf}
}

@article{sun2026c3box,
  title   = {{C3Box}: A {CLIP}-based Class-Incremental Learning Toolbox},
  author  = {Sun, Hao and Zhou, Da-Wei},
  journal = {arXiv preprint arXiv:2601.20852},
  year    = {2026}
}

@inproceedings{wu2016visualconcepts,
  title={What Value Do Explicit High Level Concepts Have in Vision to Language Problems?},
  author={Wu, Qi and Shen, Chunhua and Liu, Lingqiao and Dick, Anthony and van den Hengel, Anton},
  booktitle={Proceedings of the IEEE Conference on Computer Vision and Pattern Recognition},
  pages={203--212},
  year={2016}
}
